\documentclass[10pt,twocolumn,letterpaper]{article}

\usepackage{wacv}
\usepackage{times}
\usepackage{epsfig}
\usepackage{graphicx}
\usepackage{amsmath}
\usepackage{amssymb}
\usepackage[accsupp]{axessibility}  % Improves PDF readability for those with disabilities.
\usepackage{wacv}
\usepackage{times}
\usepackage{epsfig}
\usepackage{graphicx}
\usepackage{amsmath}
\usepackage{amssymb}
\usepackage{times}
\usepackage{epsfig}
\usepackage{graphicx}
\usepackage{amsmath}
\usepackage{amssymb}
\usepackage{multirow}
\usepackage{float}
\usepackage{bm}

\def\wacvPaperID{****} % Enter the WACV Paper ID here

\wacvfinalcopy % *** Uncomment this line for the final submission

\ifwacvfinal
\usepackage[breaklinks=true,bookmarks=false]{hyperref}
\else
\usepackage[pagebackref=true,breaklinks=true,colorlinks,bookmarks=false]{hyperref}
\fi

\begin{document}

%%%%%%%%% TITLE
\title{Single Image Deraining Network with Rain Embedding Consistency and Layered LSTM}

\author{Yizhou Li\textsuperscript{1}\quad Yusuke Monno\textsuperscript{2}\quad Masatoshi Okutomi\textsuperscript{3}\\
Tokyo Institute of Technology, Japan\\
{\tt\small \{yli\textsuperscript{1}, ymonno\textsuperscript{2}\}@ok.sc.e.titech.ac.jp, mxo@ctrl.titech.ac.jp\textsuperscript{3}}
% \and Masatoshi Okutomi
% \and
}

\maketitle

% \ifwacvfinal
% \thispagestyle{empty}
% \fi

%%%%%%%%% ABSTRACT
\begin{abstract}

Single image deraining is typically addressed as residual learning to predict the rain layer from an input rainy image.
For this purpose, an encoder-decoder network draws wide attention, where the encoder is required to encode a high-quality rain embedding which determines the performance of the subsequent decoding stage to reconstruct the rain layer. However, most of existing studies ignore the significance of rain embedding quality, thus leading to limited performance with over/under-deraining. In this paper, with our observation of the high rain layer reconstruction performance by an rain-to-rain autoencoder, we introduce the idea of ``Rain Embedding Consistency" by regarding the encoded embedding by the autoencoder as an ideal rain embedding and aim at enhancing the deraining performance by improving the consistency between the ideal rain embedding and the rain embedding derived by the encoder of the deraining network. To achieve this, a Rain Embedding Loss is applied to directly supervise the encoding process, with a Rectified Local Contrast Normalization (RLCN) as the guide that effectively extracts the candidate rain pixels. We also propose Layered LSTM for recurrent deraining and fine-grained encoder feature refinement considering different scales. Qualitative and quantitative experiments demonstrate that our proposed method outperforms previous state-of-the-art methods particularly on a real-world dataset. Our source code is available at http://www.ok.sc.e.titech.ac.jp/res/SIR/.

\end{abstract}

%%%%%%%%% BODY TEXT
\section{Introduction}
\label{sec:introduction}

Single image deraining is the task to remove the rain from a single rainy image and has been treated as a significant process, since the images captured under rainy conditions are heavily degraded, which may disturb many high-level computer vision tasks, such as object detection~\cite{objectdetection}, video surveillance~\cite{videosur}, and autonomous driving~\cite{deepdriving}. Single image deraining is very challenging since it is a highly ill-posed problem to decompose an observed rainy image into the rain layer and the rain-free background image.

Various deraining methods based on a physical or a subjective prior on rain streaks or background have been proposed~\cite{2018globalsparse,gmm,2017bilayer}. However, these prior-based methods have limited applicability for the images captured under complex rainy situations. On the other hand, data-driven learning-based deraining methods based on a Convolutional Neural Network~(CNN) have recently demonstrated their superior performance on different benchmark datasets (see~\cite{li2019single,review2021,yang2020single} for reviews). 

\begin{figure}[t!]
\begin{center}
\includegraphics[width=1.0\linewidth]{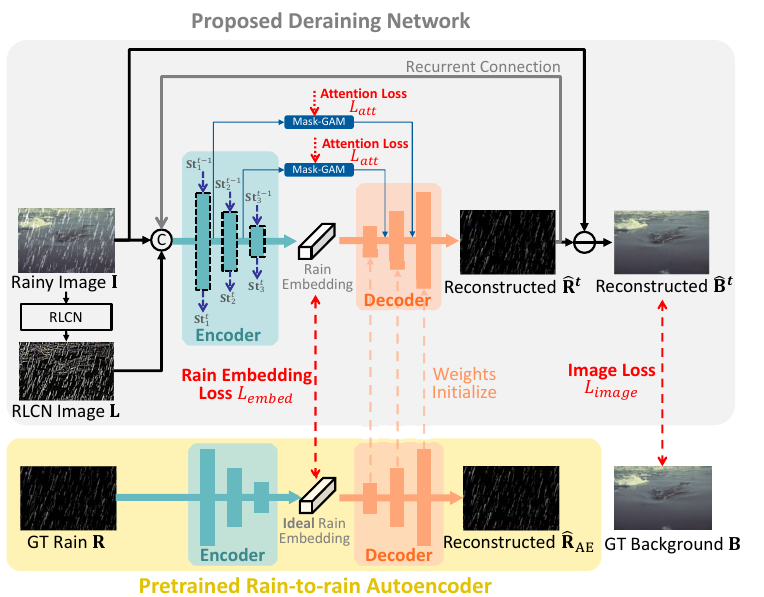}
\end{center}
    \vspace{-2mm}
   \caption{An overview of the proposed network. ${\bf St}$ denotes cross-stage states for LSTM with hidden states ${\bf H}$ and cell states ${\bf C}$.}
   \vspace{-4mm}
\label{fig:intro}
\end{figure}

%-------------------------------------------------------------------------

An encoder-decoder model is one of widely applied network architectures for single image deraining~\cite{non-local, attgan, ERL-Net, confidenceguided}, where the typical target of the encoder-decoder is to predict the residuals ${\bf R}={\bf I}-{\bf B}$ between the input rainy image ${\bf I}$ and the rain-free background image ${\bf B}$~\cite{DDN}.
Since the residuals correspond to the rain layer, the predicted rain-free image $\hat{\bf B}$ is derived by subtracting the predicted residuals $\hat{\bf R}$ from the input rainy image ${\bf I}$ as $\hat{\bf B}={\bf I}-\hat{\bf R}$. In this residual deraining encoder-decoder, the encoder is required to encode a latent high-quality rain embedding well representing the rain layer information, which is a base for high-performance rain layer generation by the subsequent decoding process by the decoder~\cite{chen2021robust}. However, the deraining performance is often limited due to the learned low-quality rain embedding as the encoder fails to decompose the rain features with the input rainy image where the rain and the background are tightly coupled. This may cause insufficient rain removal (under-deraining) or lacked textures (over-deraining).

In this paper, we introduce the idea of ``Rain Embedding Consistency." To explain this, we consider a rain-to-rain autoencoder as shown in the bottom part of Fig.~\ref{fig:intro}, whose target is to reconstruct a rain layer ${\hat{\bf R}_{AE}}$ to be as close as its input Ground-truth~(GT) rain layer ${\bf R}$. Our observation is that the autoencoder derives a superior reconstruction performance with Peak Signal-to-noise Ratio~(PSNR) larger than 50 by comparing the GT background image ${\bf B}$ and predicted background image ${\hat{\bf B}_{AE}}={\bf I}-{\hat{\bf R}_{AE}}$, implying that the autoencoder succeeds to encode an ideal rain embedding sufficiently representing the rain layer. This motivates us to improve the consistency between the ideal rain embedding by the autoencoder and the rain embedding predicted by the encoder of the deraining network.

To incorporate the Rain Embedding Consistency, we first propose to simply take a L1 loss function between the two embeddings, where we name it as Rain Embedding Loss as shown in Fig.~\ref{fig:intro}, to supervise and constraint the encoding process of the deraining network. We also propose to apply a Rectified Local Contrast Normalization~(RLCN) as an encoder input (as shown in Fig.~\ref{fig:intro}) to guide the encoding process, which extracts the pixels having a higher pixel intensity than the average pixel intensity of its neighborhood pixels. We find that, since rain streaks usually show a higher pixel intensity, the extracted pixels by RLCN cover almost all the rain candidates including weak rain streaks. Thus, RLCN can be used as a rain location prior to guide the encoder to learn better convolutional filters focusing on extracting rain features. We also propose a Mask Guided Attention Module (Mask-GAM) to replace general skip connection from encoder to decoder~\cite{unet}, which further augments the decoding phase by intensifying rain-related part of encoder features with an accurate learned rain location map before feeding to decoder.

With the above proposed components, we construct our proposed deraining network with Rain Embedding Consistency, as illustrated in Fig.~\ref{fig:intro}. To further improve the performance, we also propose a Recurrent Neural Network~(RNN) framework with Long Short-term Memory~(LSTM), which have shown great potential in single image deraining task~\cite{rescan, BRN, prenet}. Different from previous methods that apply LSTM on a single scale~\cite{BRN, prenet} of the network, we propose Layered LSTM that applies LSTM to every scale of the encoder, and let the encoded features at each scale flow to the subsequent stages as shown in Fig.~\ref{fig:intro}, where the later stages can selectively compensate the features that effectively represent the rain layer, and forget the non-related features not representing the rain layer. With this stage-by-stage and scale-by-scale fine-grained rain feature refinement using Layered LSTM, our proposed network can approach the Rain Embedding Consistency better.

Through experiments, we demonstrate that our proposed deraining network with Rain Embedding Consistency and Layered LSTM outperform state-of-the-art deraining methods by a large margin for a real-world benchmark dataset, while keeping computational efficiency. We also show the experiments on several synthetic datasets, further showing the superiority of the proposed methods, which has good resistance to over/under-deraining issues with its robust rain embedding encoding.

\section{Related Work}

We here briefly review existing single image deraining methods. Please refer to the survey papers~\cite{li2019single,review2021,yang2020single} for other classes of the methods, such as video-based methods and raindrop removal methods. 

\subsection{Prior-based Methods}

Many traditional approaches treat the single image deraining as an image decomposition problem, where a rainy image is commonly modeled as an addition of a rain layer and a rain-free background layer~\cite{2018globalsparse,2012deraining, gmm, 2017hierarchical,2017bilayer}. To regularize the decomposition problem, various priors have been proposed to model rain streaks with different directions and scales, such as Gaussian mixture model-based prior~\cite{gmm}, sparsity-based prior~\cite{2018globalsparse,2012deraining,2015deraining,2017bilayer}, and high-frequency component-based prior~\cite{2012deraining, 2017hierarchical}. A more complex and realistic image formation model, such as a screen-blend model~\cite{2015deraining}, has also been studied.

Although these traditional methods based on a physical or subjective prior demonstrate favorable performance on specific situations, they fail to remove complex rain patterns of real-world scenes where the prior does not hold.

\begin{figure*}[t!]
\begin{center}
\includegraphics[width=0.95\linewidth]{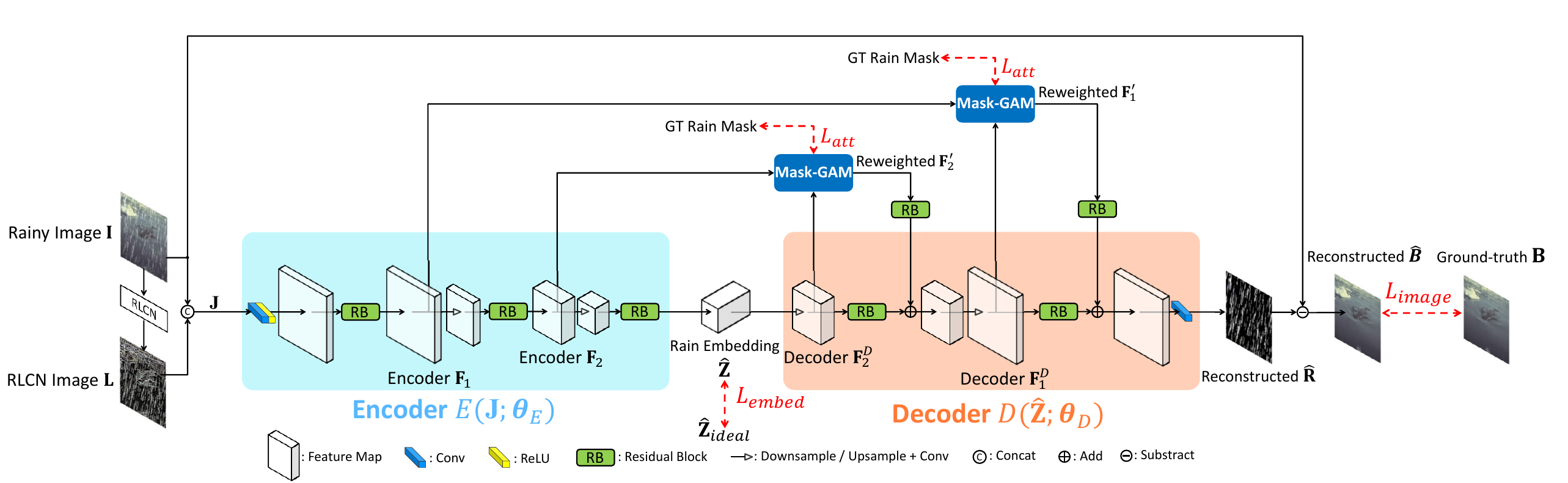}
% \includepdf[[pages={1}]]{net_overview_new_pdf.pdf}
\end{center}
    \vspace{-3mm}
   \caption{The overall architecture of proposed ECNet. It includes four scales where the number of channels for each scale is 32, 64, 128, and 256, respectively. For simplicity of illustration, we here show the three scales version of the network architecture.}
   \vspace{-3mm}
\label{fig:network}
\end{figure*}

\subsection{Learning-based Methods}

Because of the strong representation capability of CNN, CNN-based deraining methods have been extensively studied in recent years. Residual learning is commonly applied, where the network is trained to predict the residuals between the rainy and the rain-free background images, which correspond to the rain layer~\cite{DDN}. Based on the residual learning, different kinds of network architecture, such as multi-scale fusion modules~\cite{DRD-Net, MSPFN, jointagg, dcsfn, jorder-e, jorder, DID-MDN}, a density-label-guide network~\cite{DID-MDN}, a residual-guide network~\cite{resguide}, a confidence-measure-guided network~\cite{confidenceguided}, a non-locally enhanced network~\cite{non-local}, have been proposed to boost the capacity of the network. Generative adversarial networks~\cite{derainingcgan} and variational auto-encoders~\cite{VID, CVID} have also been introduced to capture the visual characteristics of the rain streaks. These methods have achieved convincing performance with their carefully designed network architectures. However, the complex deraining process is supervised with only the image-level loss function in terms of deraining output, which is difficult for the model to learn optimal latent rain features and thus limiting the deraining performance.

The methods based on an RNN is becoming a new trend to progressively remove the rain by dividing the whole deraining process into multiple stages, where LSTM~\cite{LSTM} or GRU~\cite{GRU} is applied to selectively forget noise features and refine task-related features stage-by-stage. The study~\cite{rescan} proposes to apply an RNN with GRU~\cite{GRU} and channel-wise attention to predict the rain layer and share the information across recurrent stages in feature levels. The study~\cite{prenet} applies recursive residual blocks with LSTM~\cite{LSTM} to build a simple and fast baseline for the deraining. The study~\cite{BRN} proposes the bilateral LSTM for the information sharing between the rain layer and the background layer. However, these methods apply the LSTM or GRU to a single scale of the whole deraining pipeline, without considering the potential of addressing the feature refinement on different scales separately, thus showing limited refinement performance.

\section{Proposed Method}

\subsection{Overall Architecture}

As shown in Fig.~\ref{fig:intro}, we use a rain-to-rain autoencoder for rain embedding supervision and our main deraining network, for both of which we adopt the same encoder-decoder architecture as shown in the encoder-decoder part of Fig.~\ref{fig:network}, where Residual Blocks~(RBs)~\cite{he2016deep} are used as feature extractors for both the encoder and the decoder.

To train the autoencoder, the ground-truth rain layer is generated as ${\bf{R}} = {\bf I} - {\bf B}$, where ${\bf{I}}$ and ${\bf{B}}$ are a pair of rainy image ${\bf{I}}$ and ground-truth background image ${\bf{B}}$ sampled from training data. The autoencoder takes the ground-truth rain layer as the input and encodes ideal rain embedding as $\hat{\bf Z}_{ideal}=E_{AE}({\bf{R}}; {\boldsymbol \theta}_E^{AE})$, where $E_{AE}$ represents the encoder of the autoencoder with input ${\bf{R}}$ and parameters ${\boldsymbol \theta}_E^{AE}$. The decoder then reconstructs the rain layer form the rain embedding as $\hat{\bf R}_{AE}=D_{AE}({\hat{\bf{Z}}_{ideal}}; {\boldsymbol \theta}_D^{AE})$, where $D_{AE}$ represents the decoder of the autoencoder with input ${\hat{\bf{Z}}_{ideal}}$ and parameters ${\boldsymbol \theta}_D^{AE}$. 
As the purpose of the autoencoder is to make reconstructed rain layer $\hat{\bf R}_{AE}$ to be as close as its ground-truth input ${\bf{R}}$, we simply pre-train the autoencoder using a self-supervision loss $L_{self}=||{\bf R} - \hat{\bf R}_{AE}||_2$.

Fig.~\ref{fig:network} illustrates the overall architecture of our proposed deraining network with Rain Embedding Consistency, which we call Embedding Consistency Network~(ECNet). Our target is to learn the encoder-decoder network to make the predicted rain layer $\hat{\bf R}$ that is as close as ground-truth rain layer ${\bf{R}}$, such that an accurate rain-free background prediction $\hat{\bf B}={\bf I} - \hat{\bf R}$ can be derived.

The architecture of our network is explained as follows. We first calculate the RLCN image~$\bf{L}$ from the input rainy image~$\bf{I}$ to extract the candidate pixels suggesting rain regions. The calculation of the RLCN image will be detailed in Sec.~\ref{sec:embedloss}. Then, the channel-wise concatenation of $\bf{I}$ and $\bf{L}$ is used as the network input, such that ${\bf J}=concat({\bf{I}}, {\bf{L}})$. The network encodes and derives a predicted rain embedding by ${\hat{\bf{Z}}}=E({\bf J}; {\boldsymbol \theta}_E)$, where $E$ represents the encoder with input  ${\bf J}$ and parameters ${\boldsymbol \theta}_E$.
Then, the rain layer is reconstructed from the rain embedding ${\hat{\bf{Z}}}$ by the decoder $D$ with parameters ${\boldsymbol \theta}_D$ as $\hat{\bf R}=D(\hat{\bf Z}; {\boldsymbol \theta}_D)$. Instead of applying the skip connection directly passing the encoder feature ${\bf F}$ to the decoder, we propose Mask-GAM to generate an accurate rain attention map~${\bf \hat{M}}$ to derive rain-attentive features~${\bf F'} = {\bf \hat{M}}\circ{\bf F}$, which is passed to the decoder. The details of Mask-GAM will be introduced in Sec.~\ref{sec:lcn-gam}.

The loss function to training the network will be the weighted summation of three terms as  
\begin{equation} 
    L=\lambda_{embed}L_{embed} + \lambda_{att}L_{att} + \lambda_{image}L_{image},
\end{equation}
where $\lambda_{embed}$, $\lambda_{att}$ and $\lambda_{image}$ are coefficients to balance each loss term. The first loss $L_{embed}$ is our proposed Rain Embedding Loss that uses the encoded ideal rain embedding $\hat{\bf Z}_{ideal}=E_{AE}(\bf{R})$ by the autoencoder to train the encoder with Rain Embedding Consistency.
The second loss $L_{att}$ is Attention Loss to supervise the learning of the attention map in our proposed Mask-GAM. The third loss is an Image Loss, for which we use negative SSIM loss~\cite{BRN}, which can be represented by $L_{image}=-\mathrm{SSIM}({\bf B}, \hat{\bf B})$ to evaluate the difference between the ground-truth and reconstructed background images. 

We detail each proposed component in Sec~\ref{sec:embedloss} and Sec~\ref{sec:lcn-gam} which is utilized to construct the proposed ECNet. Furthermore, we introduce our proposed Layered LSTM to further improve the performance in Sec~\ref{sec:recurrent}, where we call the resultant network ECNet+LL.

\subsection{Rain Embedding Consistency with RLCN}
\label{sec:embedloss}

As we introduced in Section~\ref{sec:introduction}, the observation of high rain layer reconstruction performance by the rain-to-rain autoencoder implies that the autoencoder can derive an ideal rain embedding $\hat{\bf Z}_{ideal}=E_{AE}({\bf R})$ by its encoder, where its decoder is able to decode and derive the rain layer $\hat{\bf R}_{AE}$ which is very close to its input $\hat{\bf R}$. This motivates us to reuse the pretrained weights ${\boldsymbol \theta}_D^{AE}$ from $D_{AE}$ to initialize the decoder $D$'s weights ${\boldsymbol \theta}_D$, which enables us to focus on training the encoder $E$ to improve the Rain Embedding Consistency between the autoencoder's ideal rain embedding ${\hat{\bf Z}_{ideal}}$ and predicted rain embedding by the encoder of the proposed deraining network ${\hat{\bf{Z}}}=E({\bf J})$.
To approach the Rain Embedding Consistency, we propose to use a feature-level Rain Embedding Loss expressed as
\begin{equation}
    L_{embed} =||\hat{\bf Z}_{ideal} - \hat{\bf{Z}}||_1,
\label{eq:embedloss}
\end{equation}
to directly supervise the encoder in terms of the feature space and force the encoder's output embedding $\hat{\bf Z}$ to approach the ideal rain bmbedding ${\hat{\bf Z}_{ideal}}$.

To further regularize the rain embedding learning, we proposed to apply RLCN to the input rainy image as 
\begin{equation} 
    {\bf L}(i,j,c)=\frac{\phi_{rec}\left({\bf I}_c(i,j)-\mu_{{\bf I}_c}(i,j)\right)}{\sigma_{{\bf I}_c}(i,j)+\epsilon},
\label{eq:lcn}
\end{equation}
where ${\bf L}$ is the output RLCN image, $(i, j)$ denotes the pixel coordinate, $c \in \left(R,G,B\right)$ represents the color channel, ${\bf I}_c$ is c-th channel of the input rainy image, $\mu_{\bf I}(i, j)$ denotes the mean pixel intensity within a local square window centered at the pixel $(i, j)$, $\sigma_{\bf I}(i, j)$ denotes the standard deviation of the pixel intensities within the local window, $\epsilon$ is a small value for numerical stability, and $\phi_{rec}(\cdot)=\mathrm{max}(\cdot, 0)$ is the rectification function that outputs the max value compared with zero.

\begin{figure}[t!]
\begin{center}
\includegraphics[width=1.0\linewidth]{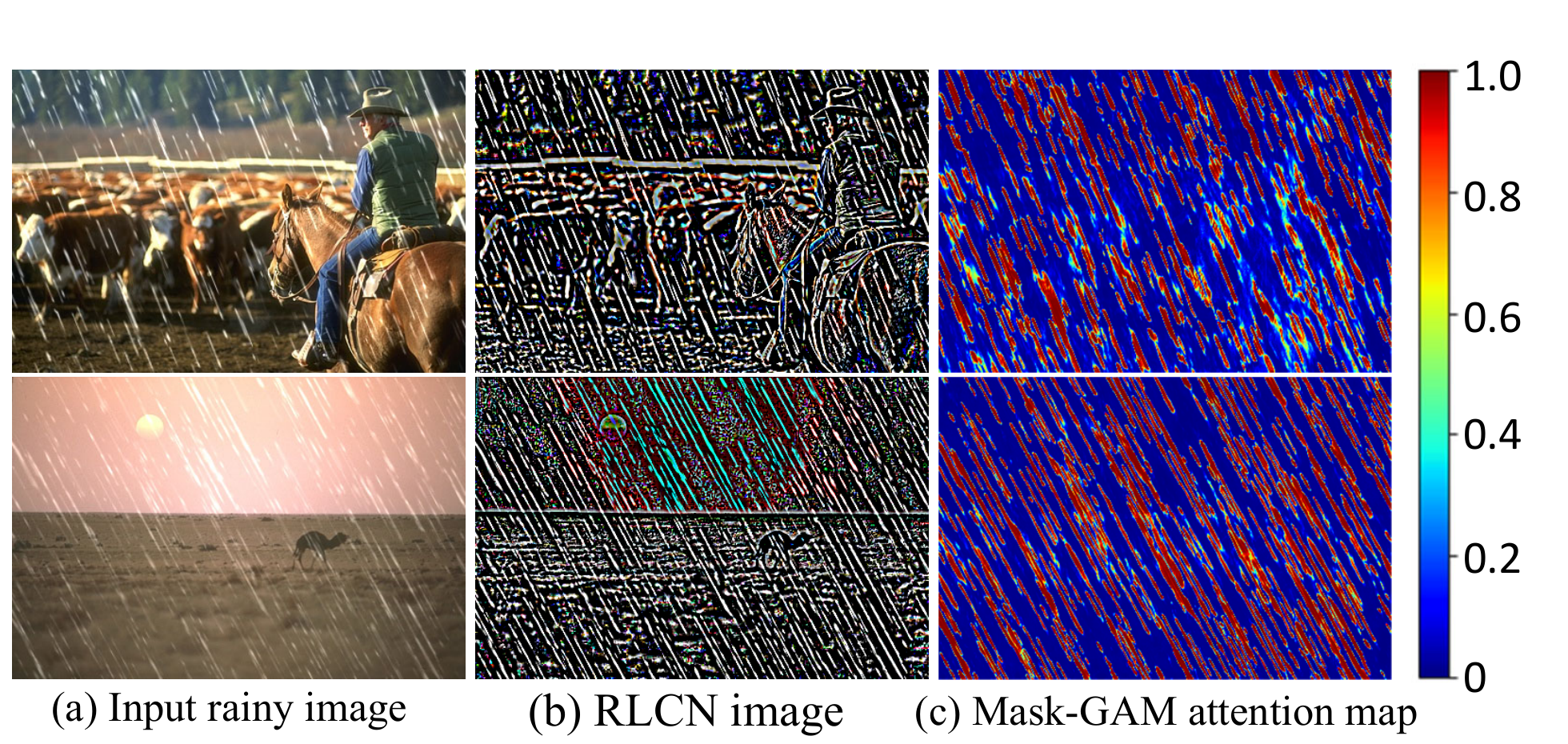}
\end{center}
    \vspace{-4mm}
   \caption{Two examples of (a)~input rainy images, (b)~our proposed RLCN images, (c)~our proposed Mask-GAM attention maps.}
    \vspace{-4mm}
\label{fig:lcn}
\end{figure}

In Eq.~(\ref{eq:lcn}), the subtracted value, ${\bf I}(i,j)-\mu_{\bf I}(i,j)$, takes a high absolute value if the pixel intensity of the pixel $(i,j)$ is significantly higher or lower than the intensities of the other pixels within the local window. Because we focus on the rain, which usually shows a higher pixel intensity, the max function is applied to perform the rectification and filter out negative values. The positive value after the rectification is further normalized by the local standard deviation $\sigma_{\bf I}(i, j)$, which corresponds to local contrast, to enable the extraction of the pixels for weak rain regions as well as strong rain regions. As show in Fig.~\ref{fig:lcn}, the derived RLCN image can cover most of the rain regions and thus can be used as a guide for the rain embedding encoding, which helps the encoder to learn rain-focus convolutional filters, thus improving the Rain Embedding Consistency performance by the encoder. 

\begin{figure}[t!]
\begin{center}
\includegraphics[width=0.95\linewidth]{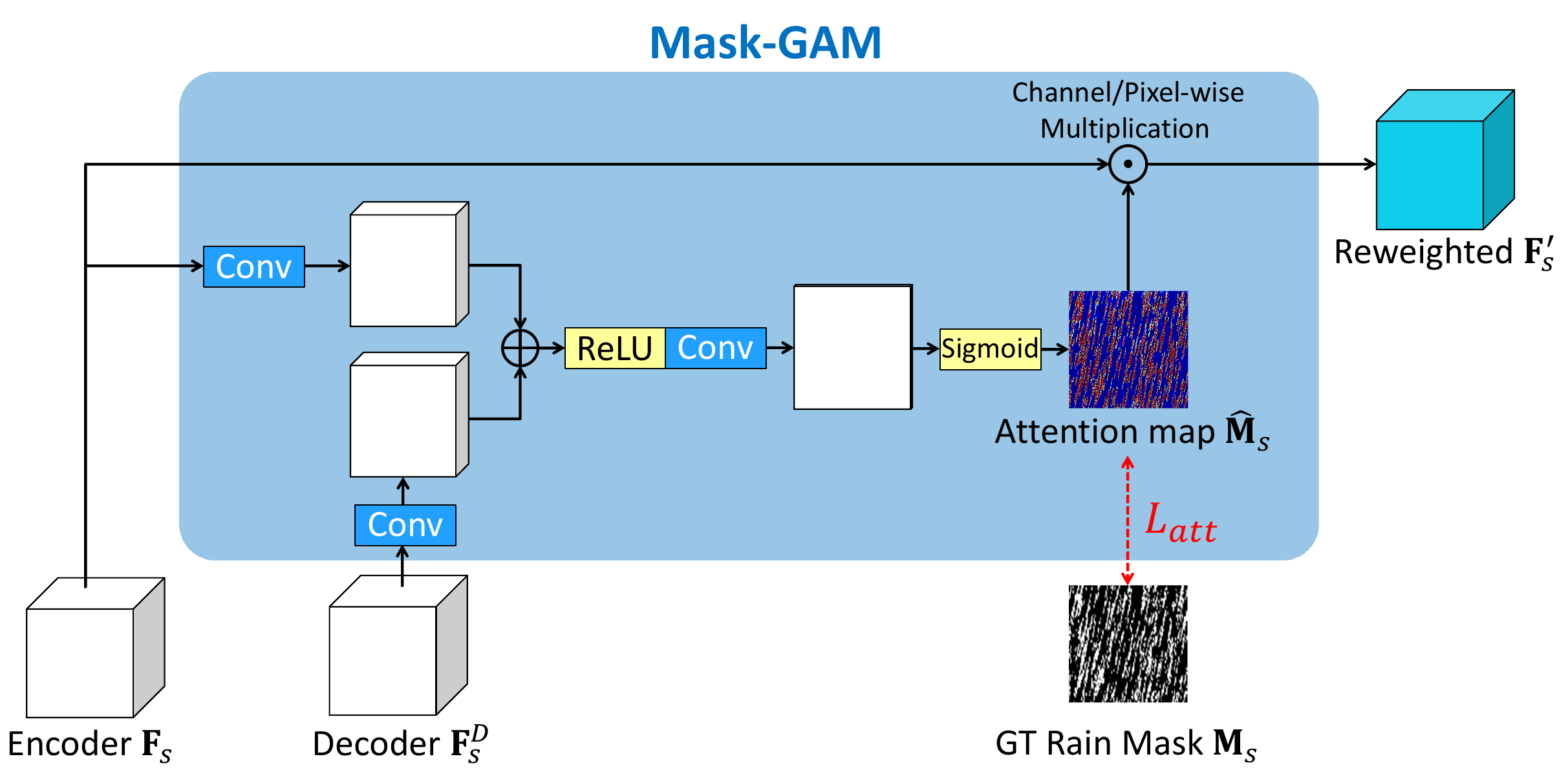}
% \includepdf[[pages={1}]]{net_overview_new_pdf.pdf}
\end{center}
    \vspace{-3mm}
   \caption{Architecture of proposed Mask-GAM.}
   \vspace{-4mm}
\label{fig:att}
\end{figure}

\subsection{Mask Guided Attention Module (Mask-GAM)}
\label{sec:lcn-gam}

\begin{figure*}[t!]
\begin{center}
\includegraphics[width=0.95\linewidth]{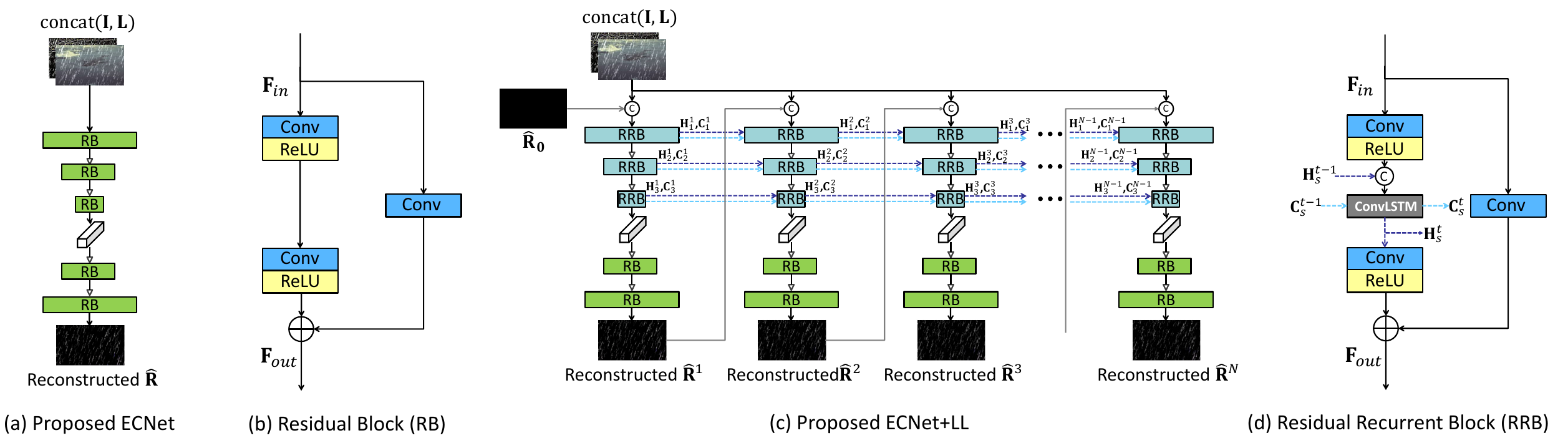}
\end{center}
    \vspace{-4mm}
    \caption{Our proposed recurrent framework. For simplicity the Mask-GAM is ignored in this illustration. Based on proposed ECNet~(a) which uses RBs~(b), we construct ECNet+LL~(c) by introducing the recurrent framework with Layered LSTM that replaces RBs in the encoder with the proposed RRBs~(d).}
    \vspace{-4mm}
\label{fig:recurrent}
\end{figure*}

Skip connection is widely applied in the encoder-decoder to reuse the encoder features to prevent the over-smoothing of the decoding. However, as RLCN image contains some additional edges regarding the non-rain regions as shown in Fig.~\ref{fig:lcn}, there exist unexpected features non-related to rain included in the encoder features. Therefore, passing the features directly from the encoder to the decoder could disturb the decoding phase. To address this issue, Mask-GAM of Fig.~\ref{fig:att} is proposed to generate a pixel-level attention map to more accurately regard the rain location as shown in Fig.~\ref{fig:lcn}. The attention map is used to reweight the encoder features to remove the unexpected features and highlight the rain-related features.

Following the mathematical symbol notations in Fig.~\ref{fig:att}, Mask-GAM is expressed as
\begin{equation}
    \hat{\bf M}_s =\phi_2({\bf W}^A_s * (\phi_1({\bf W}_{s} * {\bf{F}}_{s} + {\bf{W}}_{s}^{D} * {\bf F}_{s}^D))),
\label{eq:attention}
\end{equation}
where $\hat{\bf M}_s$ is the attention map for scale $s$ of the encoder-decoder and ${\bf F}_{s}$ and ${\bf F}_{s}^{D}$ represent the features from the encoder and the decoder, respectively. ${\bf W}_{s}$ and ${\bf W}_{s}^{D}$ denote the weights for $1\times1$ convolutional layers with $n$ filters, where we set $n$ as the half of the channel number of ${\bf{F}}_{s}$, which is equal to the channel number of ${\bf{F}}_{s}^D$. The convolution operation, which is expressed as $*$, is performed to ${\bf{F}}_{s}$ and ${\bf{F}}_{s}^D$, respectively, and the resultant features are combined by the element-wise summation. Then, the Rectified Linear Unit~(ReLU) \cite{glorot2011deep} function~$\phi_{1}$ is applied to the combined features, which followed by a one-channel $1\times1$ convolutional layer with the weight~${\bf W}_s^A$ to sum up the $n$-channel feature maps. The sigmoid activation function $\sigma_{2}$ is then applied to derive the attention map within the range of $[0, 1]$. Finally, using the obtained attention map, the re-weighted features ${\bf F}'_s$ is derived by ${\bf F}'_s = \hat{\bf M}_s \circ {\bf F}_s$ as rain-attentive features and then sent to the decoder, where $\circ$~denotes the channel and pixel-wise multiplication. Then, to supervise the learning of the attention map, we use the ground-truth rain mask ${\bf M}$ derived by thresholding the ground-truth rain layer ${\bf R}$ to construct a loss function as
\begin{equation} 
    L_{att}=\frac{\sum_{s=1}^{S}||\hat{\bf M}_s - {\bf M}||_2}{S},
\label{eq:gamloss}
\end{equation}
where $S$ is the scale number of the encoder-decoder and $s$ is the scale index. This attention module not only excludes the non-related features from the encoded features, but also can be seen as a tiny additional branch of the proposed encoder-decoder network whose output target is the rain regions, thus forcing the network to derive features that can represent the information inside rain regions.

\subsection{Recurrent Framework with Layered LSTM}
\label{sec:recurrent}

In this section, we introduce the proposed ECNet+LL network which is built based on the proposed ECNet. Firstly, we reformulate the ECNet network of Fig.~\ref{fig:recurrent}(a) into a recurrent style as shown in Fig.~\ref{fig:recurrent}(c), where the output of the decoder corresponds to the rain layer $\hat{\bf R}^t$ of current stage $t$. Then, the current output $\hat{\bf R}^t$ is used as the input of the next stage $t+1$ and the overall network flow is repeated until the iteration reaches the maximum recurrent time $N$. Here we set $\hat{\bf R}_0={\bf 0}$. For every recurrent stage, the network parameters are shared and optimized during the training phase.

Then, we introduce a Residual Recurrent Block~(RRB) with LSTM to each scale of the encoder as shown in Fig.~\ref{fig:recurrent}(c). As they let the features flow across successive recurrent stages with a scale-by-scale style, we name this architecture as Layered LSTM~(LL). As shown in Fig.~\ref{fig:recurrent}(d), a convolutional LSTM layer is added inside RB of Fig.~\ref{fig:recurrent}(b) to construct the RRB. As the information flow of LSTM, the hidden and cell state output of stage $t-1$, which are denoted as ${\bf H}^{t-1}_s$ and ${\bf C}^{t-1}_s$ respectively with $s$ describing the scale index, and the features from the last ReLU layer are sent to the LSTM layer, where the gating signal inside LSTM decides which part of the information is preserved or thrown away. Then, the LSTM outputs the hidden and cell state of current stage $t$, which is denoted as ${\bf H}^{t}_s$ and ${\bf C}^{t}_s$, where ${\bf H}^{t}_s$ is used as the input of the second convolutional layer and both of ${\bf H}^{t}_s$ and ${\bf C}^{t}_s$ are sent to the RRB of the next stage $t+1$. By applying the recurrent framework with Layered LSTM, the network is able to conduct a scale-by-scale fine-grained feature refinement for encoding, where the subsequent stages of encoding are able to retain and refine the rain-related features in terms of each scale separately, and forget the features non-related to rain, thus boosting the network to improve Rain Embedding Consistency.

As for the loss function after introducing the recurrent framework, we calculate the loss $L$ for each stage $t$ separately, which is denoted as $L_t$, and recursive supervision is introduced by adding them together as the final loss function. This can be represented as 
\begin{equation} 
    \vspace{-2mm}
    L_{recur}=\sum_{t=1}^{N}\lambda_tL_t
    \vspace{-0.5mm}
\end{equation}
where $\lambda_t$ is used to balance the loss of each stage $t$. 

%------------------------------------------------------------------------
\section{Experimental Results}

\subsection{Implementation Details}

Our ECNet and ECNet+LL networks were implemented using Pytorch~\cite{paszke2019pytorch} and trained using one NVIDIA Tesla P100 GPU. For all the experiments, the networks share the same training settings. The size of the local window for the RLCN image calculation is set to 9$\times$9 and the recurrent time $N$ of the proposed network is set to 6. We set $\lambda_{embed}=0.02$, $\lambda_{att}=0.1$, $\lambda_{image}=1$ to balance the influence of each loss, and the $\lambda_{t}$ is set to 0.5 for $t\leq 5$, and 1.5 for final stage $t=6$. The patch size of 96$\times$96 is used for the training, with the batch size set to 16. ReLU~\cite{glorot2011deep} is used as the activation function and the gradient clipping of 5 is applied to stabilize the network training. Adam optimizer~\cite{adam} is used to train the network with an initial learning rate $1e^{-3}$. We trained the network for 100 epochs while decaying the learning rate when reaching 25, 50, 75 epochs by multiplying 0.2.

\subsection{Study of Proposed Components}

The studies to confirm the effectiveness of each proposed component was conducted using real-world SPA-Data dataset~\cite{spanet}. SPA-Data contains high-quality 638,492 training images and 1,000 test images generated using video redundancy and human supervision. For time efficiency, we randomly selected 6,385 images (1\%) from the original training data to train the networks and tested on the full 1,000 test images.

\vspace{1mm}
\noindent
% \subsubsection{Effects of Rain Feature Consistency and RLCN}
\textbf{Effects of Rain Embedding Consistency with RLCN:} To explore the effectiveness of the proposed method with Rain Embedding Consistency with the guide of the RLCN image, the following four networks are evaluated in Table~\ref{tb:ablation_representation}. ED is the basic encoder-decoder that applies only the encoder-decoder part of the proposed network and uses only a rainy image as the network input, and it is trained using only Image Loss $L_{image}$. ED + EmbedLoss means Rain Embedding Consistency is considered by training the network with Rain Embedding Loss $L_{embed}$. ED + RLCN means that the RLCN image is used as the input for the encoder. ED + EmbedLoss + RLCN means that both the Rain Embedding Loss and the RLCN guide are applied. Table~\ref{tb:ablation_representation} shows that the training with $L_{embed}$ actually benefit the network performance and also the RLCN guide remarkably improves the network performance. When combining $L_{embed}$ and the RLCN guide, the network achieves the best results. Hereafter, we use ED + EmbedLoss + RLCN as the base of the proposed method (Proposed-Base) for the next experiment.

\begin{table}[t!]
\caption{Effects of Rain Embedding Consistency with RLCN.}
\vspace{-1mm}
\label{tb:ablation_representation}
\begin{center}
\small
\renewcommand\tabcolsep{3pt}
\renewcommand\arraystretch{1.1}
\begin{tabular}{l c c}
\hline
Methods & PSNR & SSIM\\
\hline %\cmidrule(lr){2-4}
ED & 40.01 & 0.978\\
\hline %\cmidrule(lr){2-4}
ED + EmbedLoss & 40.38 & 0.979\\
\hline %\cmidrule(lr){2-4}
ED + RLCN & 40.73 & 0.980 \\
\hline %\cmidrule(lr){2-4}
ED + EmbedLoss + RLCN (Proposed-Base) & 40.85 & 0.980\\

\hline
\end{tabular}
\end{center}
\vspace{-5mm}
\end{table}

\vspace{1mm}
\noindent
% \subsubsection{Effects of Mask-GAM}
\textbf{Effects of Mask-GAM:} To explore the effectiveness of the proposed Mask-GAM, Proposed-Base and and the following three networks are evaluated in Table~\ref{tb:ablation_maskgam}. Proposed-Base + Skip applies the direct skip connection between the encoder and the decoder. Proposed-Base + Self-attention means that self-attention~\cite{oktay2018attention} is utilized to reweight the encoder features before feeding to the decoder. Proposed-Base + Mask-GAM means that our proposed Mask-GAM with the Attention Loss $L_{att}$ is applied to supervise the attention learning and this method corresponds to the proposed ECNet.  Table~\ref{tb:ablation_maskgam} shows that the direct skip connection provides limited improvements as it may introduce the features non-related to rain from the encoder to the decoder. With self-attention, this phenomenon is relieved to some extent. And with the proposed Mask-GAM, the network achieves the best performance as it explicitly guides the learning of the attention and leaves rain-related features only.

\begin{table}[t!]
\caption{Effects of Mask Guided Attention Module.}
\vspace{1mm}
\label{tb:ablation_maskgam}
\begin{center}
\small
\renewcommand\tabcolsep{6pt}
\renewcommand\arraystretch{1.1}
\begin{tabular}{l c c}
\hline
Methods & PSNR & SSIM\\
\hline %\cmidrule(lr){2-4}
Proposed-Base & 40.85 & 0.980\\
\hline %\cmidrule(lr){2-4}
Proposed-Base + Skip & 41.35 & 0.986\\
\hline %\cmidrule(lr){2-4}
Proposed-Base + Self-attention & 41.58 & 0.987\\
\hline %\cmidrule(lr){2-4}
Proposed-Base + Mask-GAM (ECNet) & 41.83 & 0.987\\

\hline
\end{tabular}
\end{center}
\vspace{-5mm}
\end{table}

\begin{table}[t!]
\caption{Effects of Recurrent Framework with Layered LSTM.}
\vspace{-3mm}
\label{tb:ablation_recur}
\begin{center}
\small
\renewcommand\tabcolsep{6pt}
\renewcommand\arraystretch{1.1}
\begin{tabular}{l c c}
\hline
Methods & PSNR & SSIM\\
\hline %\cmidrule(lr){2-4}
ECNet & 41.83 & 0.987\\
\hline %\cmidrule(lr){2-4}
ECNet + LSTM & 41.90 & 0.987\\
\hline %\cmidrule(lr){2-4}
ECNet + LayeredLSTM (ECNet+LL) & 42.49 & 0.988\\

\hline
\end{tabular}
\end{center}
\vspace{-8mm}
\end{table}

\vspace{1mm}
\noindent
% \subsubsection{Effects of Stage-wise Recurrent Framework}
\textbf{Effects of Recurrent Framework with Layered LSTM:} To explore the effectiveness of the proposed Layered LSTM, proposed ECNet and the following two networks are evaluated in Table~\ref{tb:ablation_recur}. ECNet + LSTM denotes that the network is first formulated to a recurrent style, and only the RB of the last scale of the encoder is changed to RRB, as performed in~\cite{recur-unet}. ECNet + LayeredLSTM further replaces the RBs in every scale of the encoder with RRBs, denoting the proposed Layered LSTM. This method corresponds to the full version of our method, i.e., ECNet+LL. As we can see in Table~\ref{tb:ablation_recur}, only applying LSTM to the last scale of the encoder almost does not benefit the network performance, while the proposed Layered LSTM shows substantial performance boost as LSTM is applied to every scale of the encoder to conduct fine-grained rain feature refinement scale-by-scale.

\begin{figure*}[t!]
\begin{center}
\includegraphics[width=0.65\linewidth]{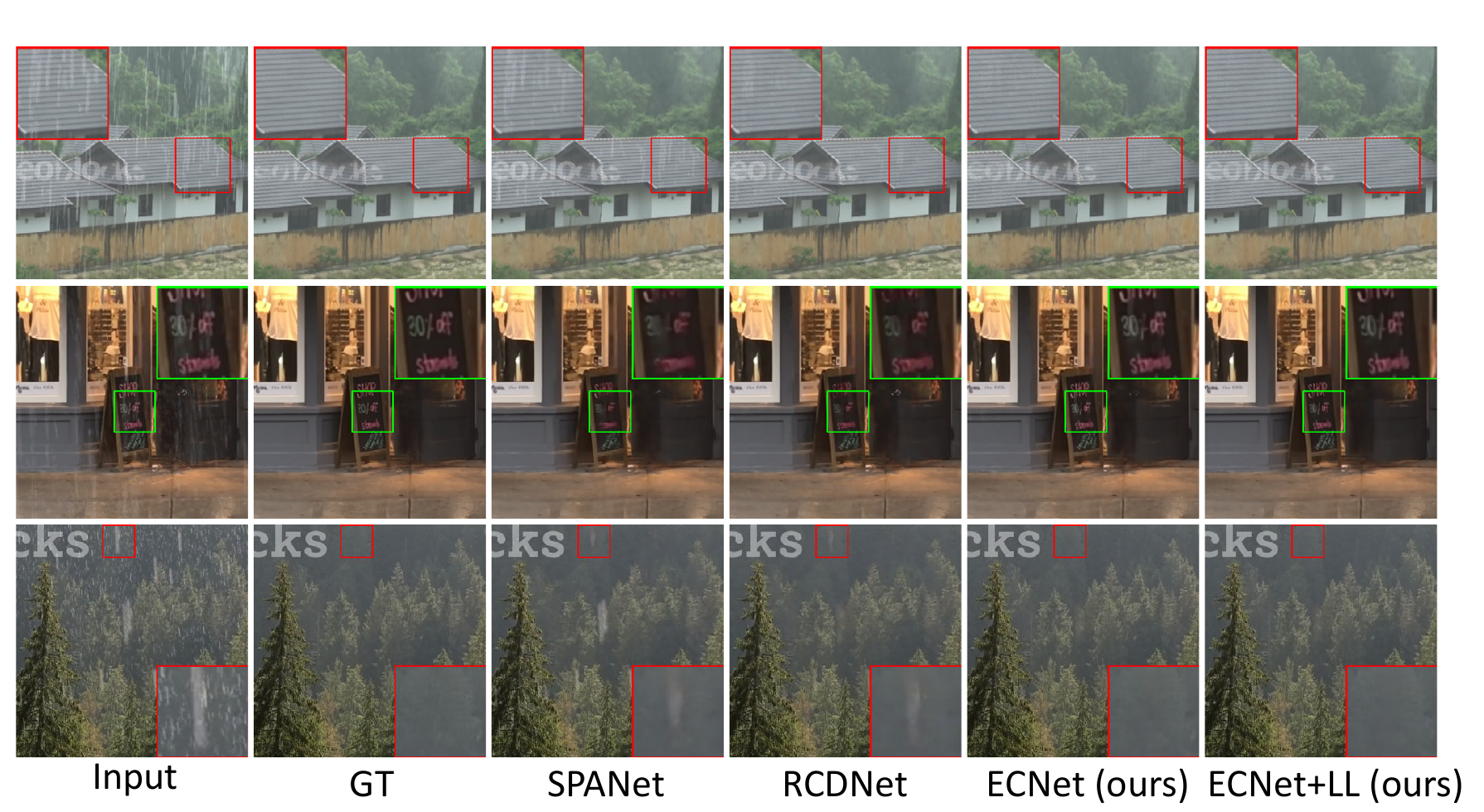}
% \fbox{\rule{0pt}{3in} \rule{.9\linewidth}{0pt}}
\end{center}
    \vspace{-4mm}
   \caption{Qualitative comparison on real-world SPA-Data}
   \vspace{-2mm}
\label{fig:spa}
\end{figure*}

\subsection{Comparison on Real-world Dataset}

\noindent
\textbf{Compared methods:}
We evaluate the deraining performance of proposed ECNet and ECNet+LL for real-world images of SPA-Data~\cite{spanet}. We trained the proposed networks twice using the full training set (638,492 images) and 1\% of the training set (6,385 images, to show the learning performance with a small amount of data), respectively. Since the full dataset is very large and retraining takes a huge amount of time, we chose two previous state-of-the-art methods, SPANet~\cite{spanet} and RCDNet~\cite{RCD-Net}, for comparison, which provide a public pretrained model on the full training set. SPANet is the method that proposes SPA-Data and RCDNet is a recent state-of-the-art method. For the 1\% training set, we retrained the models for SPANet and RCDNet as no pretrained model is provided for the 1\% set and retraining can be done with reasonable training time. 

\begin{table}[t!]
\caption{Quantitative comparison on real-world SPA-Data (Red: rank 1st; Blue: rank 2nd) with different amount of training data from SPA-Data.}
\vspace{-2mm}
\label{tb:psnr_spa}
\begin{center}
\footnotesize
\renewcommand\tabcolsep{1.5pt}
\renewcommand\arraystretch{1.3}
\begin{tabular}{c c c c | c c | c c | c c}
\hline
\multicolumn{2}{c|}{Methods} & \multicolumn{2}{c|}{SPANet~\cite{spanet}} & \multicolumn{2}{c|}{RCDNet~\cite{RCD-Net}} & \multicolumn{2}{c|}{ECNet (ours)} & \multicolumn{2}{c}{ECNet+LL (ours)}\\
\cline{2-10}
\multicolumn{2}{c|}{Metrics} & PSNR & SSIM & PSNR & SSIM & PSNR & SSIM & PSNR & SSIM\\
\hline %\cmidrule(lr){2-4}
\multicolumn{2}{c|}{6385 (1\%)} & 36.37 & 0.971 & 39.75 & 0.982 & \textcolor{blue}{\textbf{41.83}} & \textcolor{blue}{\textbf{0.987}} & \textcolor{red}{\textbf{42.49}} & \textcolor{red}{\textbf{0.988}}\\
\multicolumn{2}{c|}{Full} & 40.04 & 0.984 & 41.05 & 0.985 & \textcolor{blue}{\textbf{43.62}} & \textcolor{blue}{\textbf{0.990}} & \textcolor{red}{\textbf{44.32}} & \textcolor{red}{\textbf{0.991}}\\
\hline
\end{tabular}
\end{center}
\vspace{-7mm}
\end{table}

\begin{table*}[t!]
\caption{Quantitative comparison on synthetic datasets 
(Red: rank 1st; Blue: rank 2nd)}
\label{tb:psnr}
\begin{center}
\renewcommand\tabcolsep{3pt}
\renewcommand\arraystretch{1.0}
\begin{tabular}{c c c c | c c | c c | c c | c c | c c}
\hline
\multicolumn{2}{c|}{Datasets} & \multicolumn{2}{c|}{Rain100H} & \multicolumn{2}{c|}{Rain200H} & \multicolumn{2}{c|}{Rain100L} & \multicolumn{2}{c|}{Rain200L} & \multicolumn{2}{c|}{Rain800} & \multicolumn{2}{c}{Average}\\
\cline{2-14}
\multicolumn{2}{c|}{Metrics} & PSNR & SSIM & PSNR & SSIM & PSNR & SSIM & PSNR & SSIM & PSNR & SSIM & PSNR & SSIM\\
\hline %\cmidrule(lr){2-4}
% \multicolumn{2}{c|}{DDN~\cite{DDN}} & 26.79 & 0.814 & 26.10 & 0.807 & 34.61 & 0.959 & 34.39 & 0.960 & 25.47 & 0.836\\
\multicolumn{2}{c|}{RESCAN~\cite{rescan}} & 28.82 & 0.867 & 27.95 & 0.862 & 38.09 & 0.980 & 38.43 & 0.982 & 28.36 & 0.872 & 32.33 & 0.913\\
\multicolumn{2}{c|}{SIRR~\cite{SIRR}} & 22.03 & 0.714 & 22.17 & 0.726 & 32.31 & 0.926 & 32.21 & 0.931 & 22.73 & 0.762 & 26.29 & 0.812\\
\multicolumn{2}{c|}{PReNet~\cite{prenet}} & 30.31 & 0.910 & 29.47 & 0.907 & 37.21 & 0.978 & 37.93 & 0.983 & 26.82 & 0.888 & 32.35 & 0.933\\
\multicolumn{2}{c|}{JORDER-E~\cite{jorder-e}} & 30.22 & 0.898 & 29.23 & 0.894 & 39.36 & 0.985 & 39.13 & 0.985 & 27.92 & 0.883 & 33.17 & 0.929 \\
\multicolumn{2}{c|}{RCDNet~\cite{RCD-Net}} & 31.26 & 0.912 & 30.18 & 0.909 & \textcolor{red}{\textbf{39.76}} & \textcolor{red}{\textbf{0.986}} & \textcolor{blue}{\textbf{39.49}} & \textcolor{blue}{\textbf{0.986}} & 28.66 & 0.893 & \textcolor{blue}{\textbf{33.87}} & 0.937\\
\multicolumn{2}{c|}{BRN~\cite{BRN}} & \textcolor{blue}{\textbf{31.32}} & \textcolor{red}{\textbf{0.924}} & \textcolor{red}{\textbf{30.27}} & \textcolor{red}{\textbf{0.919}} & 38.16 & 0.982 & 38.86 & 0.985 & 28.31 & 0.896 & 33.38 & \textcolor{blue}{\textbf{0.941}}\\
\multicolumn{2}{c|}{ECNet (ours)} & 29.80 & 0.903 & 28.54 & 0.893 & 38.21 & 0.981 & 38.37 & 0.983 & \textcolor{blue}{\textbf{28.80}} & \textcolor{blue}{\textbf{0.901}} & 32.74 & 0.932\\
\multicolumn{2}{c|}{ECNet+LL (ours)} & \textcolor{red}{\textbf{31.43}} & \textcolor{blue}{\textbf{0.921}} & \textcolor{blue}{\textbf{30.22}} & \textcolor{blue}{\textbf{0.912}} & \textcolor{blue}{\textbf{39.66}} & \textcolor{red}{\textbf{0.986}} & \textcolor{red}{\textbf{39.72}} & \textcolor{red}{\textbf{0.987}} & \textcolor{red}{\textbf{29.26}} & \textcolor{red}{\textbf{0.905}} & \textcolor{red}{\textbf{34.06}} & \textcolor{red}{\textbf{0.942}}\\
\hline
\end{tabular}
\end{center}
\vspace{-5mm}
\end{table*}

\vspace{1mm}
\noindent
\textbf{Results:}
Table~\ref{tb:psnr_spa} shows the PSNR and the SSIM results. We can see that proposed ECNet and ECNet+LL provide remarkably better results for real-world rainy images, showing the strong applicability to real-world situations. Also, both of proposed methods derives better performance even when they are trained on only 1\% of the training data, compared with SPANet and RCDNet trained on the full training data, showing that the proposed methods considering Rain Embedding Consistency can achieve high performance with only a small amount of training data. Fig.~\ref{fig:spa} shows the qualitative results of each method trained on the full training set. For samples on the first and the third rows, both of proposed ECNet and ECNet+LL can remove the rain more completely compared with the other methods. For samples on the second row, our methods remove the rain effectively and succeed to maintain the background textures of a black board, while SPANet and RCDNet fails to reproduce them. These results validate that proposed ECNet and ECNet+LL are able to better deal with varied rain conditions containing different intensities and shapes that generally appear in real-world scenarios, and can address under/over-deraining issues more robustly. More deraining samples are shown in supplementary material.

\subsection{Comparison on Synthetic Datasets}
\label{sec:synthetic}

\noindent
\textbf{Compared methods:}
We evaluate proposed ECNet and ECNet+LL on five synthetic benchmark datasets, Rain100H~\cite{jorder}, Rain200H~\cite{jorder}, Rain100L~\cite{jorder}, Rain200L~\cite{jorder}, Rain800~\cite{derainingcgan}. Rain100H/200H synthesize the heavy rain conditions, Rain100L/200L synthesize the light rain conditions, and Rain800 synthesizes both heavy and light rainy images. We compare the proposed networks with state-of-the-art methods including semi-supervised SIRR~\cite{SIRR}, rain-mask-based~JORDER-E~\cite{jorder-e}, model-driven RCDNet~\cite{RCD-Net}, and RNN-based RESCAN~\cite{rescan}, PReNet~\cite{prenet}, and BRN~\cite{BRN}. RCDNet and BRN are recent state-of-the-art methods. We used the pretrained models of the above methods for Rain100H, Rain200H and Rain100L, and retrain models for Rain200L and Rain800 under the original settings of each method as no pretrained model is provided except BRN for Rain200L.

\begin{figure}[t!]
\begin{center}
\includegraphics[width=0.95\linewidth]{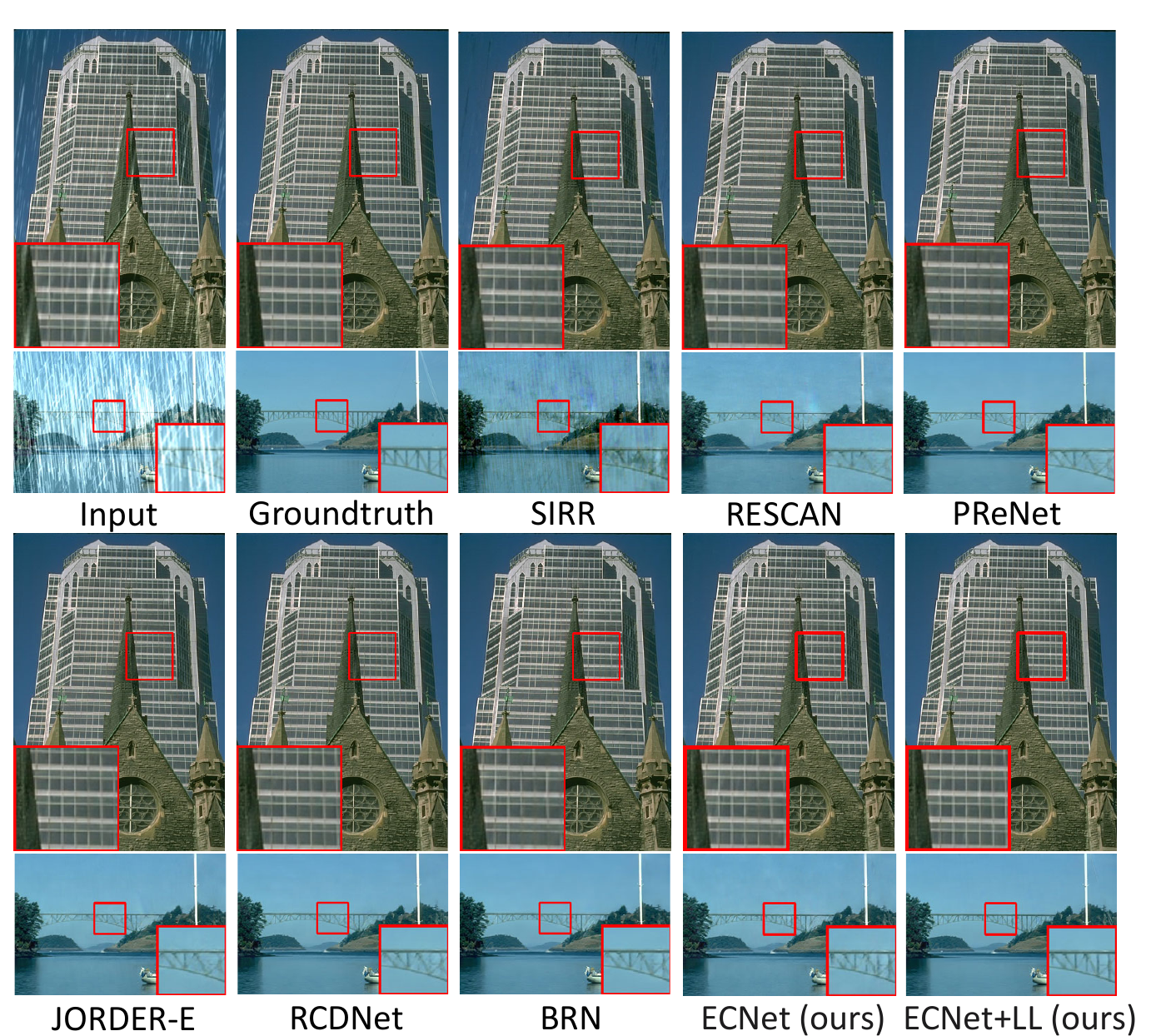}
% \fbox{\rule{0pt}{3in} \rule{.9\linewidth}{0pt}}
\end{center}
    \vspace{-4mm}
   \caption{Qualitative comparison on synthetic datasets}
   \vspace{-5mm}
\label{fig:rain100hl}
\end{figure}

\vspace{1mm}
\noindent
\textbf{Results:}
Table~\ref{tb:psnr} reports the PSNR and the SSIM results, where proposed ECNet+LL provides the best average performance compared with the other methods, showing its strong adaptability under different rain conditions. Proposed ECNet also derives fairly good performance considering its light-weightness as will be discussed in Section~\ref{sec:efficiency}. Fig.~\ref{fig:rain100hl} shows two qualitative samples on the Rain100L dataset and the Rain100H dataset. For the samples on the first row, we can see that both of ECNet and ECNet+LL can maintain the window frames of the building that have similar appearance to the rain streaks, while other methods show apparent over-deraining as they also remove background textures. Similarly, for the heavy rain samples on the second row, both of our methods reproduce the bridge textures more effectively. These results show the strength of our methods to deal with under/over-deraining issues.

\begin{figure}[t!]
\begin{center}
\includegraphics[width=1.0\linewidth]{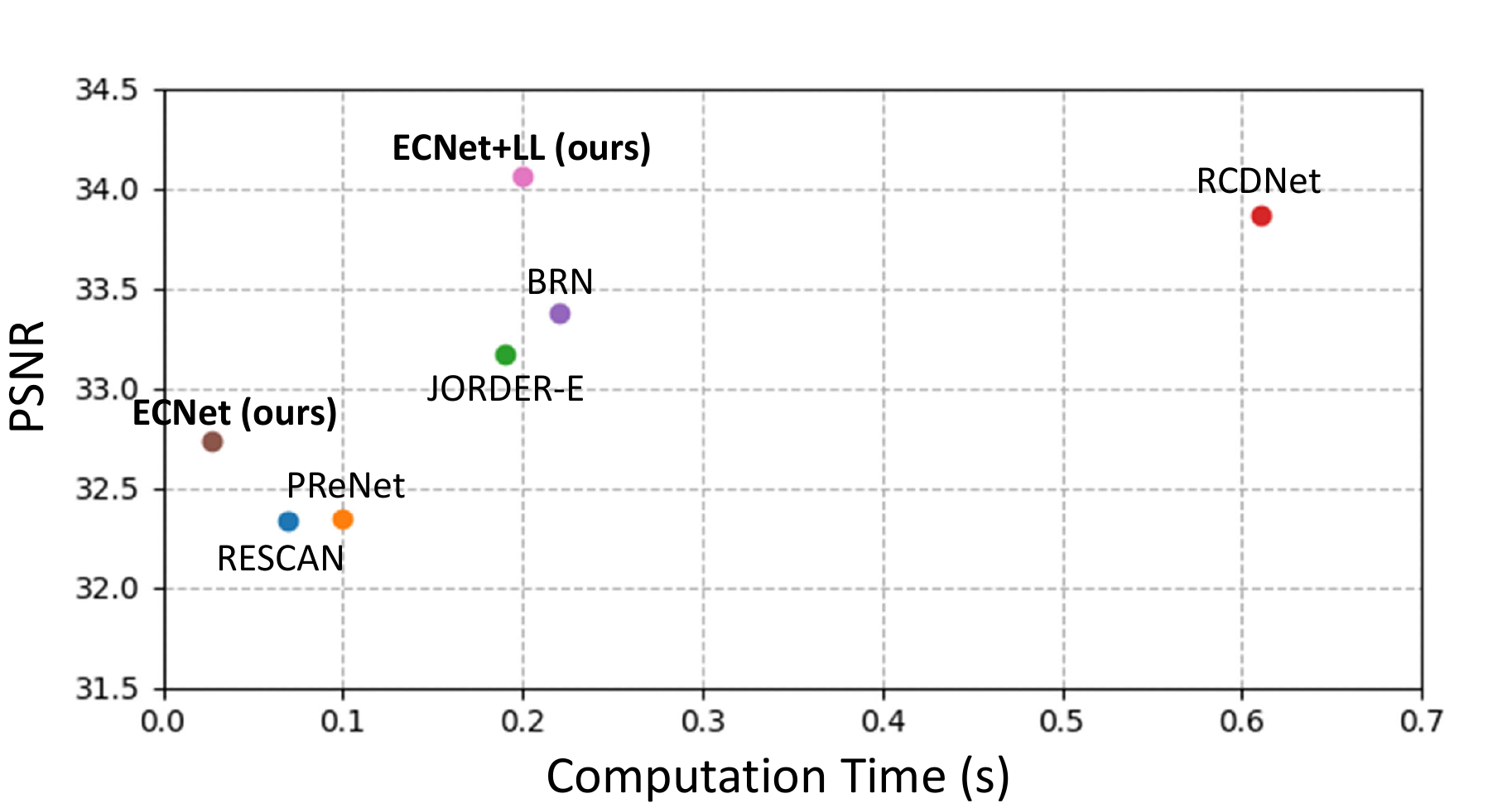}
\end{center}
   \vspace{-3mm}
   \caption{Computational time vs. PSNR plot for average PSNR on all the five synthetic datasets}
   \vspace{-7mm}
\label{fig:time}
\end{figure}

\subsection{Computational Time}
\label{sec:efficiency}

Figure~\ref{fig:time} shows the computational time versus PSNR plot for average PSNR on all the five synthetic datasets (except SIRR because of its lower PSNR). The running time is recorded with one NVIDIA P100 GPU for the image whose resolution is $480\times320$. As shown in Fig.~\ref{fig:time}, proposed ECNet+LL achieves the best PSNR result with almost the same computation time as BRN and JORDER-E and much less computational time than RCDNet, showing the better performance-speed balance compared with other competing methods. Also, proposed ECNet shows the best performance among the methods with computation time that is no more than 0.1s. Proposed ECNet can achieve a real-time inference with 37FPS (0.027s per inference) on a NVIDIA P100 GPU, which could be meaningful to real-world applications where computational time is an important factor.

\section{Conclusion}

In this paper, we have proposed a single image deraining network considering Rain Embedding Consistency with an ideal rain embedding encoded by the rain-to-rain autoencoder to address the difficulty to learn a better rain embedding. First, we have introduced the idea of Rain Embedding Consistency and train the network with a Rain Embedding Loss to incorporate the Rain Embedding Consistency, with an RLCN image as the guide for the encoder to learn better rain-focused convolutional filters. Next, we have proposed Mask-GAM with ground-truth rain mask supervision to reweight the encoder features to leave only rain-attentive features which are passed to the decoder. Finally, we have proposed a recurrent framework with Layered LSTM to form the encoding into a recurrent process with LSTM on each scale of the encoder to iteratively refine the encoded rain features with a fine-grained scale-by-scale style. The quantitative and qualitative evaluations using representative deraining benchmark datasets have demonstrated that our proposed method outperforms existing state-of-the-art deraining methods, where it is particularly noteworthy that our method clearly achieves the best performance on a real-world dataset.
%\footnote{We plan to make the source code publicly available.}

\newpage
{\small
\bibliographystyle{ieee_fullname}
\bibliography{egbib}
}

\end{document}